%% file: root.tex
\pdfoutput=1
\documentclass[letterpaper, 10 pt, conference]{ieeeconf}

\IEEEoverridecommandlockouts
\usepackage{tabularx,amsmath,amsfonts,siunitx,pifont,amssymb,svg}
\usepackage{xcolor, bm, float, multirow, courier, mathtools, url}
\usepackage{subfigure}
\usepackage[ruled,vlined]{algorithm2e} 
\usepackage{balance}

\makeatletter
\let\NAT@parse\undefined
\makeatother
\usepackage[hidelinks]{hyperref}  % for href
\hypersetup{
  colorlinks   = true, % Colours links instead of ugly boxes
  urlcolor     = blue, % Colour for external hyperlinks
  linkcolor    = blue, % Colour of internal links
  citecolor    = red   % Colour of citations
}

\usepackage{scalerel}
\usepackage{tikz}
\usetikzlibrary{svg.path}

\definecolor{orcidlogocol}{HTML}{A6CE39}
\tikzset{
	orcidlogo/.pic={
		\fill[orcidlogocol] svg{M256,128c0,70.7-57.3,128-128,128C57.3,256,0,198.7,0,128C0,57.3,57.3,0,128,0C198.7,0,256,57.3,256,128z};
		\fill[white] svg{M86.3,186.2H70.9V79.1h15.4v48.4V186.2z}
		svg{M108.9,79.1h41.6c39.6,0,57,28.3,57,53.6c0,27.5-21.5,53.6-56.8,53.6h-41.8V79.1z M124.3,172.4h24.5c34.9,0,42.9-26.5,42.9-39.7c0-21.5-13.7-39.7-43.7-39.7h-23.7V172.4z}
		svg{M88.7,56.8c0,5.5-4.5,10.1-10.1,10.1c-5.6,0-10.1-4.6-10.1-10.1c0-5.6,4.5-10.1,10.1-10.1C84.2,46.7,88.7,51.3,88.7,56.8z};
	}
}

\newcommand\orcidicon[1]{\href{https://orcid.org/#1}{\mbox{\scalerel*{
				\begin{tikzpicture}[yscale=-1,transform shape]
				\pic{orcidlogo};
				\end{tikzpicture}
			}{|}}}}

\newcommand{\ea}{\textit{et al}.~}
\newcommand{\ie}{\textit{i}.\textit{e}.}

\usepackage{pdfpages}

\title{\LARGE \bf
    Receding-Horizon Perceptive Trajectory Optimization for \\
    Dynamic Legged Locomotion with Learned Initialization
} 

\author{Oliwier Melon \orcidicon{0000-0001-6092-9477}, 
    	Romeo Orsolino \orcidicon{0000-0001-9847-2601}, 
    	David Surovik \orcidicon{0000-0002-9454-5874}, 
		Mathieu Geisert \orcidicon{0000-0002-5651-8736}, \\
		Ioannis Havoutis \orcidicon{0000-0002-4371-4623} 
		and Maurice Fallon \orcidicon{0000-0003-2940-0879}%

\thanks{This work was supported by the UKRI/EPSRC RAIN Hub
[EP/R026084/1] and the
EU H2020 Projects MEMMO and THING, the EPSRC grant `Robust Legged Locomotion' [EP/S002383/1] and a Royal Society University Research Fellowship (Fallon). This work was conducted as part of ANYmal Research, a community to advance legged robotics.
The authors are with Oxford Robotics
Institute, University of Oxford, UK. Email:
{\tt\small \{omelon, rorsolino, dsurovik, mathieu, ioannis, mfallon\}@robots.ox.ac.uk}.}	
}

\makeatletter
\def\endthegraphy{%
    \def\@noitemerr{\@latex@warning{Empty `thebibliography' environment}}%
    \endlist
}
\makeatother

\begin{document}
	
\setlength{\abovedisplayskip}{4pt}
\setlength{\belowdisplayskip}{4pt}
	
\maketitle 

\input{notation.tex}

\begin{abstract}
\input{abstract.tex}
\end{abstract}
	
\input{introduction.tex}
\input{related_works.tex}
\input{framework.tex}
\input{results.tex}
\input{conclusion.tex}

\balance
\bibliographystyle{IEEEtran}
\bibliography{library}
    
\end{document}

%% file: notation.tex
\newcommand{\iH}{{Heuristic}}
\newcommand{\iL}{{LMTR}}

% Additional features
\newcommand{\statep}{\boldsymbol\chi} % state in the State Prediction section

% LMTR
\newcommand{\task}{{\mathbf x}}
\newcommand{\traj}{\boldsymbol \tau}
\newcommand{\mode}{{\mathbf z}}
\newcommand{\latent}{{\mathbf z}}
\newcommand{\goal}{\boldsymbol\gamma}
\newcommand{\terr}{\boldsymbol\eta}
\newcommand{\terrline}{\mathbf o}
\newcommand{\state}{\boldsymbol\chi} % state in the LMTR section

\newcommand{\stathoriz}[1]{#1^\prime}

% Optimization
\newcommand{\primals}{\traj}

%% file: abstract.tex
To dynamically traverse challenging terrain, legged robots need to continually perceive and reason about upcoming features, adjust the locations and timings of future footfalls and leverage momentum strategically. 
We present a pipeline that enables flexibly-parametrized trajectories for perceptive and dynamic quadruped locomotion to be optimized in an online, receding-horizon manner.
The initial guess passed to the optimizer affects the computation needed to achieve convergence and the quality of the solution. 
We consider two methods for generating good guesses.
The first is a \emph{heuristic} initializer which provides a simple guess and requires significant optimization but is nonetheless suitable for adaptation to upcoming terrain. 
We demonstrate experiments using the ANYmal C quadruped, with fully onboard sensing and computation, to cross obstacles at moderate speeds using this technique.
Our second approach uses \emph{latent-mode trajectory regression} (\iL{}) to imitate expert data%
---while avoiding invalid interpolations between distinct behaviors---%
such that minimal optimization is needed.
This enables high-speed motions that make more expansive use of the robot's capabilities.
We demonstrate it on flat ground with the real robot and provide numerical trials that progress toward deployment on terrain.
These results illustrate a paradigm for advancing beyond short-horizon dynamic reactions, toward the type of intuitive and adaptive locomotion planning exhibited by animals and humans.

%% file: introduction.tex
\section{Introduction} \label{sec:introduction}

State-of-the-art legged-locomotion controllers can maintain dynamic motion while handling perturbations from external forces on moderately uneven terrain.
Alternatively, locomotion planners can reason over longer horizons to strategically plan footholds and paths across more prominent obstacles, such as those seen in industrial settings.
However, the benefits of these two paradigms---speed and foresight, respectively---are not trivial to combine.
As a result legged robots have limited performance in scenarios that demand careful foothold selection and momentum shaping.

Meanwhile, legged animals can dynamically traverse obstacles with relative ease.
They are able to make rapid inferences about their circumstances
and subconsciously relate them to relevant prior experiences
to leverage the properties of their bodies in a coordinated manner.
Biological locomotion serves as inspiration to motion-generation research in robotics.

\subsection{Planning Dynamic Motion Online} \label{sec:intro_trajectory_optimization}
Dynamically complex robot motion plans are commonly generated through trajectory optimization%
---a process which minimizes an objective function while satisfying equality and inequality constraints.
The Trajectory Optimization for Walking Robots (TOWR) framework~\cite{winklerGaitTrajectoryOptimization2018} introduced a solution parametrization appropriate for handling the discontinuities inherent to legged locomotion, which ordinarily creates problems for gradient descent, without imposing overly rigid assumptions.
Adapting this approach for online use carries several additional challenges, such as the need to minimize the required computation time and the need to employ the optimizer in a \emph{receding-horizon} manner without compromising the solution's expressiveness.

One increasingly popular strategy for online planning is to produce high-quality initial guesses by leveraging offline experience, 
so that only minor computation is needed to adapt to a scenario seen at runtime.
Many earlier examples of this approach focused on manipulators~\cite{jetchevFast2013}, leading to later studies on more dynamic systems~\cite{mansardUsingMemoryMotion2018,lembonoLearningHowWalk2020,melonReliableTrajectoriesDynamic2020a}.
Recent works have considered more complex variants that may be needed when the relevant solution space is multimodal~\cite{lynchLearningLatentPlans2020,surovikLearningExpertSkillSpace}.
For the most part, these works impose strong assumptions about the beginning and end of the motion plan being produced.
For generalized, sustained behavior by dynamic quadrupeds, a formulation is needed that can handle their multimodality---and to do so without being required to routinely revisit a small and restrictive class of checkpoint states.

\begin{figure}[t]
	\includegraphics[width=\columnwidth]{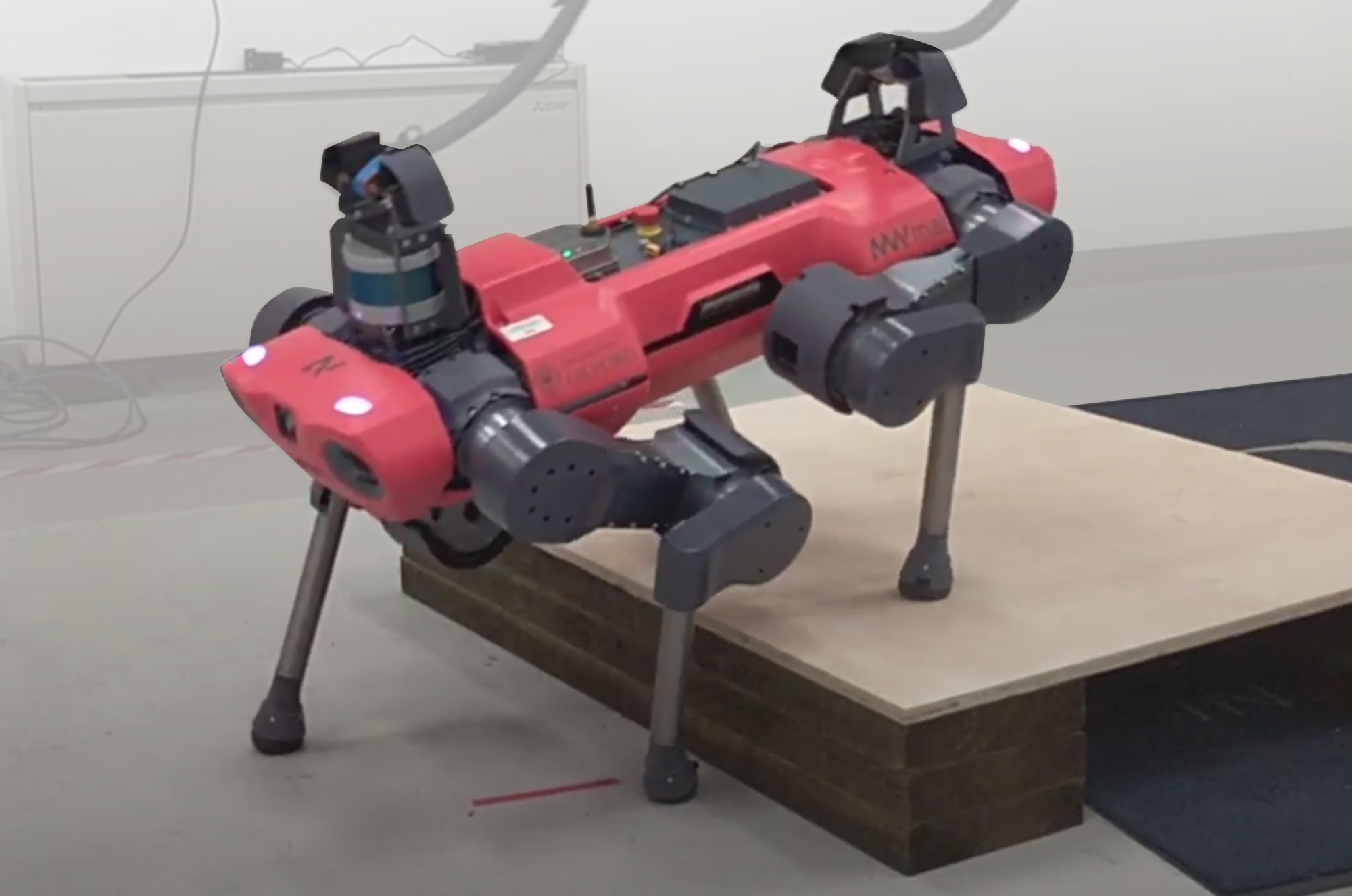}
	\caption{
        The ANYmal C quadruped using online, receding-horizon trajectory optimization and real-time perception to 
        cross a \SI{0.20}{\metre} step
        as demonstrated in the accompanying \href{https://youtu.be/Qqs5hhp3hHQ}{video}.
    }
    \vspace{-1em}
	\label{fig:main}
\end{figure}

\begin{figure} [t]
    \vspace{0.7em}
	\includegraphics[width=\columnwidth]{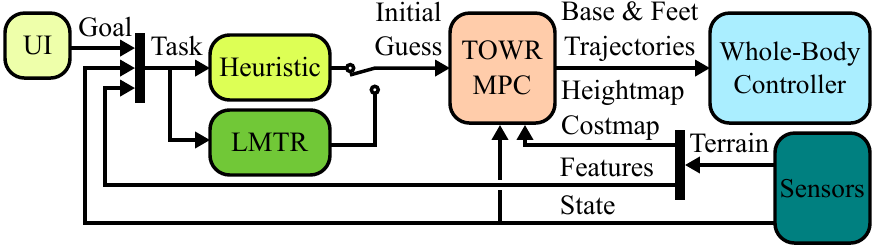}
	\vspace{-1.5em}
    \caption{
        Overview of the proposed onboard system.
        The user, or a high-level planning algorithm, specifies the goal as a SE(2) base pose within a couple meters of the robot.
        The sensors provide the estimated state, along with environment features.
        Together, these form a task that is input to one of the proposed initializers, Heuristic or LMTR,
        which produces an asynchronously-parametrized trajectory segment
        for the trajectory optimizer (TOWR MPC) to refine so that it satisfies dynamic and kinematic constraints.
        It uses the heightmap, a dense 2.5D elevation map produced using depth cameras or LIDAR, 
        and the costmap to plan favorable footholds.
        The resulting base and feet trajectories are tracked by the whole-body controller.
	}
	\label{fig:system}
    \vspace{-1em}
\end{figure}

\subsection{Contributions and Outline}
Motivated by current literature discussed in Sec.~\ref{sec:related_works}, we aim to enable \emph{online replanning} of dynamic motions that can be \emph{sustained indefinitely} while anticipating upcoming terrain features by employing \emph{diverse modes of behavior}.
This paper presents a substantial extension and integration of our previous works~\cite{melonReliableTrajectoriesDynamic2020a, surovikLearningExpertSkillSpace};
its novel contributions are:
\begin{itemize}
	\item 
    Implementation of an \emph{asynchronous}, \emph{segmentation}-based trajectory optimization formulation for legged robots based on the single rigid body dynamic (SRBD) model to enable \emph{receding-horizon} planning
    (Sec.~\ref{sec:problem_formulation}).

    \item 
    Incorporation of two alternate methods for initializing the nonlinear programming problem (NLP): 
    a heuristic that makes efficient re-use of the previous plan (Sec.~\ref{sec:heuristic_initialization}), 
    and the data-driven approach of latent-mode trajectory regression (\iL{})
    \cite{surovikLearningExpertSkillSpace}
    that can generalize the gait in anticipation of upcoming terrain (Sec.~\ref{sec:framework_lmtr}).
	
	\item 
	Evaluation of the proposed system in simulation and on hardware
	using a real ANYmal C quadruped (Sec.~\ref{sec:results}). 
\end{itemize}

Figure~\ref{fig:system} provides an overview of our replanning framework, which is detailed throughout Sec.~\ref{sec:framework}.
Additional features include future state prediction based tracking error,
state update within the solver at each solve iteration,
a contact regain behavior 
and use of real-time perception of terrain to select favorable footholds (Sec.~\ref{sec:state_prediction}--\ref{sec:whole_body_control}).
Following discussion of the findings in Sec.~\ref{sec:results}, we summarize our conclusions and identify future directions in Sec.~\ref{sec:conclusion}.

%% file: related_works.tex
\section{Related Work} \label{sec:related_works}

\subsection{Motion Planning via Model Predictive Control}
Many of the most successful approaches to dynamic legged locomotion use trajectory optimization.
They may include the kinematics and dynamics of the limbs~\cite{DaiWholeBody2014, carpentierMulticontact2018, mastalliCrocoddylEfficientVersatile2020} or use simplified models such as centroidal dynamics~\cite{OrinCentroidal2013} or single rigid body dynamics (SRBD)~\cite{winklerGaitTrajectoryOptimization2018}; they can also adopt different strategies for contact scheduling such as completely predefining the timings, only predefining the sequence~\cite{PontonOnTime2018} or fully discovering both~\cite{PosaDirectMethod2014, MordatchDiscovery2012}. However, few of these methods are suitable for online motion generation.

To be applicable online, approaches usually rely on simplified dynamics criteria like the Zero-Moment-Point~\cite{VukoBratovicZeroMoment2004}, modeling as a single rigid body with fixed orientation~\cite{DiCarloDynamic2018, bledtImplementingRegularizedPredictive2019}.
Some works decouple the perceptive foothold selection to reduce computation time~\cite{fankhauserRobustRoughTerrainLocomotion2018a, jeneltenPerceptiveLocomotionRough2020, villarrealMPCbasedControllerTerrain2020a,kimVisionAidedDynamic2020a}.
While simultaneous optimization of footholds has been achieved \cite{herdtWalkingThinkingIt2010,naveauReactiveWalkingPattern2017, winklerOnlineWalkingMotion2017},
doing it on terrain remains a challenge. 

More sophisticated nonlinear MPC approaches use the whole-body model of the legged robot but have optimization horizons that are too short to be able to adequately traverse large obstacles \cite{Farshidian2017}. Other implementations based on the full robot model employ differential dynamic programming \cite{mastalliCrocoddylEfficientVersatile2020} with an efficient feasibility-driven formulation that promises to enable high-frequency receding-horizon control. 
Another family of approaches is mixed-integer programming (MIP)~\cite{apgarFastOnlineTrajectory2018, Aceituno2018} which, however, does not scale well with a large number of decision variables. Other authors successfully used the SRBD which represents a compromise between model accuracy and complexity that allows online replanning~\cite{cebeOnlineDynamicTrajectory2020}.
While the SRBD assumes the inertia to be invariant, the changing property can be accounted and compensated for without increasing model complexity~\cite{villarrealMPCbasedControllerTerrain2020a}.

\subsection{Data-Driven Locomotion}
Model-based approaches to motion generation may be limited by optimization time or the discovery effort needed to construct complex or subtle behavior online.
This has lead to data-driven approaches that \emph{amortize} this cost and effort by conducting it offline.
Bledt \ea observed correlations between tasks and optimized behaviors to identify heuristic costs for regularizing online optimization~\cite{bledtImplementingRegularizedPredictive2019,bledtExtractingLeggedLocomotion2020}.
Alternately, machine learning approaches reduce the role of humans in curating the amortized knowledge.

A hierarchy of two reinforcement learning~(RL) policies was used in~\cite{tsounisDeepGaitPlanningControl2020a}---one for planning footholds and phase durations, and one for base and end-effector motion.
To gradually build up the complexity of the learned behaviors, Xie \ea \cite{xieALLSTEPSCurriculumdrivenLearning2020} used curriculum learning for bipedal stepping-stone motions, evaluated in simulation.
However, these methods become sample-inefficient when dealing with a large state-space and typically require considerable reward shaping.

For producing longer-horizon actions, the work of Mansard \ea \cite{mansardUsingMemoryMotion2018} ``warm-started'' nonlinear MPC using a model learned from a sampling-based approach.
Similar use of a function approximator has enabled a real quadruped to quickly plan several-steps-long dynamic trajectories which it successfully executed to ascend terrain~\cite{melonReliableTrajectoriesDynamic2020a}.
Guided Trajectory Learning uses consensus optimization to learn a function approximator that only reconstructs feasible motions~\cite{duburcqOnlineTrajectoryPlanning2020}.
In~\cite{lembonoLearningHowWalk2020}, a ``memory of motion'' of dynamic, collision-free locomotion has been generated using the \emph{HPP Loco3D} planner~\cite{tonneauEfficientAcyclicContact2018} for the \emph{Crocoddyl} solver~\cite{mastalliCrocoddylEfficientVersatile2020}.

To adequately capture highly varied behavior, multiple regressors can be used, as in computer animation of dogs and humans~\cite{zhangModeadaptiveNeuralNetworks2018, starkeLocalMotionPhases2020}.
For legged robots, \cite{cariusMPCNetFirstPrinciples2020} used
imitation learning of optimal control demonstrations, improving sample efficiency over RL.
To remove the need for multiple networks and discrete categories of actions, latent variables can be used to express continuous modal variation~\cite{lynchLearningLatentPlans2020,surovikLearningExpertSkillSpace}.

%% file: framework.tex
\section{Framework} \label{sec:framework}

In this section we present our approach to online replanning and execution of dynamic locomotion.
The developed framework is shown in Fig.~\ref{fig:system}.
We consider two alternate initialization methods. 
The \iH{} is the traditional MPC-style initialization scheme which reuses a part of the previous solution.
The remaining section of the new initial guess, for which no prior data is available, is set up using linear interpolation.
By contrast, \iL{} leverages offline experience to adapt to the current combination of state, goal, and terrain, \ie, the task.
The \iH{} imposes more conservative behavior but aptly demonstrates the receding-horizon formulation.
Meanwhile, more aggressive and adaptive motions can be expressed via \iL{}, although its deployability depends upon the quality of its training data.

\input{framework_formulation.tex}

\input{framework_heuristic.tex}
\input{framework_lmtr.tex}
\input{framework_integration.tex}

%% file: framework_formulation.tex
\subsection{Optimization Formulation} \label{sec:problem_formulation}

Our formulation is based on Hermite-Simpson direct collocation
which represents the state using cubic polynomials of Hermite form, 
\ie, the splines are defined by the values~\(\mathcal{N}\) and derivatives~\(\mathcal{\dot{N}}\) of the nodes.
We use phase-based end-effector parametrization \cite{winklerGaitTrajectoryOptimization2018},
which enables the base and end-effector nodes to be asynchronous,
allowing the phase durations to be optimized
to achieve an aperiodic contact sequence.
To enable MPC and facilitate its initialization,
we keep the dimensionality of the problem constant
by proposing a \emph{segmentation}-based formulation
with \emph{asynchronous} base and feet time horizons
as shown in Fig.~\ref{fig:segmentation}.
It is also vital for the training of the LMTR (Sec.~\ref{sec:framework_lmtr});
the regressor has a fixed-size output of a given set of optimization variables \(\primals\) representing a trajectory with a predefined time horizon.

The size of the proposed problem is: 
\begin{equation}
\begin{aligned}
	\#\primals 
	&= (3+3+3+3) \times n^\text{base}_{\text{nodes}}                                                &&\text{base-motion} \\
	&+ n_{\text{feet}} \times (n_{\text{st}} - 1 + \ n_{\text{sw}})               &&\text{durations} \\
	&+ n_{\text{feet}} \times [3 \times n_{\text{st}} + (3+2) \times {n_{\text{sw}}}] &&\text{foot-motion} \\
	&+ n_{\text{feet}} \times (3+3) \times n^\text{force/foot/st}_{\text{nodes}} \times n_{\text{st}}                         &&\text{forces} \\
\end{aligned}
\end{equation}
where the individual components are:
\subsubsection{Base-motion}
Base linear position \(\mathbf{r}(t) \in \mathbb{R}^3\), linear velocity \(\mathbf{\dot{r}}(t) \in \mathbb{R}^3\), angular position \(\bm{\theta}(t) \in \mathbb{R}^3\) and angular velocity \(\bm{\omega}(t) \in \mathbb{R}^3\) values are multiplied by the number of base nodes \(n^\text{base}_{\text{nodes}}\) which is calculated using the durations of the horizon \(\Delta T_{\text{H}}\) and the time step of the base polynomials \(dt_{\text{base-poly}}\) as~\(n_{\text{base-nodes}} = 1+\text{round}(\Delta T_{\text{H}}/dt_{\text{base-poly}})\).
\subsubsection{Durations}
For a quadruped, the number of legs \(n_{\text{feet}}\) is \num{4}.
The chosen number of stance \(n_{\text{st}}\) and swing \(n_{\text{sw}}\) phases for each foot \(i\) is \num{2} and \num{1}, respectively.
\subsubsection{Foot-motion}
The stance phase is defined by the linear position \(\mathbf{p}_{i}(t)\in \mathbb{R}^3\) values, while the swing phase has both linear position \(\mathbf{p}_{i}(t)\in \mathbb{R}^3\) and linear velocity \(\mathbf{\dot{p}}_{i}(t)\in \mathbb{R}^2\) values.
The velocity in the z-dimension is an implicit constraint set to 0.
\subsubsection{Forces}
The force values and derivatives \(\mathbf{f}_{i}(t), \mathbf{\dot{f}}_{i}(t) \in \mathbb{R}^3\) are multiplied by the number of force nodes per foot per stance \(n^\text{force/foot/st}_{\text{nodes}}\), which is \num{3} as forces are constrainted to \(\mathbf{0}\) at the start and end of each \emph{swing} phase.
For every mid-motion stance phase, this number further decreases to \num{2} as the forces at the start and end are \(\mathbf{0}\).

The problem is defined over a duration \(\Delta T\) between time \(t_{\text{0}}\) of the earliest beginning of the \emph{first} set of stances \(\Delta T_{i, \text{st1}}\) and the time \(t_{\text{f}}\) of latest ending of the \emph{second} set of stances \(\Delta T_{i, \text{st2}}\); in between, the swing phases \(\Delta T_{i, \text{sw}}\) occur.
If a leg is in swing at $t_{0,H}$, that swing is not re-optimized.
Instead, the next stance-swing-stance sequence is considered in the optimization.

\begin{figure} [t!]
	\vspace{0.3em}
	\includegraphics[width=\columnwidth]{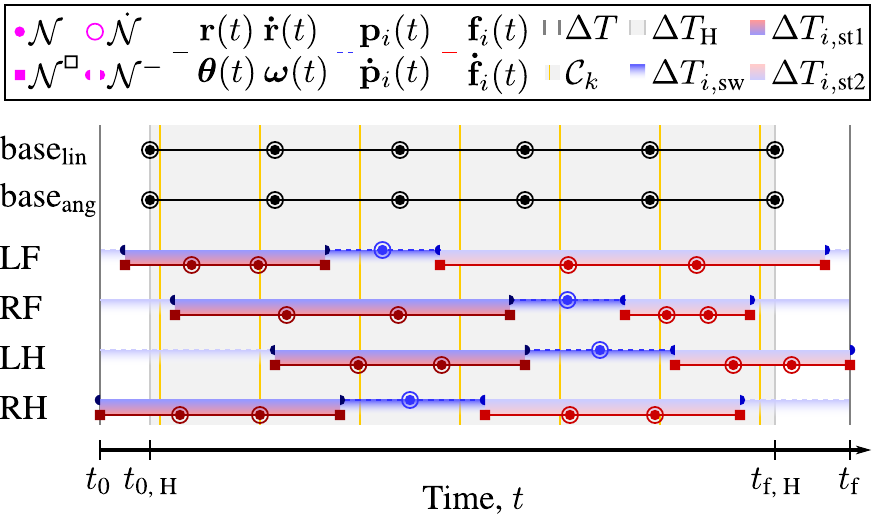}
	\caption{
		Our direct collocation, \emph{segmentation}-based problem formulation
		uses phase-based end-effector (LF, RF, LH, RH) parametrization~\cite{winklerGaitTrajectoryOptimization2018}.
		To facilitate initialization of the solver
		and allow the contact phase durations to be optimized,
		we keep the dimensionality constant
		by defining base and feet segments \emph{asynchronously}.
		The components of the figure are thoroughly described in Sec.~\ref{sec:problem_formulation}.
	}
    \vspace{-1em}
	\label{fig:segmentation}
\end{figure}

Figure~\ref{fig:segmentation} illustrates the problem formulation for the base horizon \(\Delta T_{\text{H}} = 5\,dt_{\text{base-poly}}\) which is defined with respect to the first and last base nodes occuring at \(t_{\text{0, H}}\) and \(t_{\text{f, H}}\), respectively.
The figure shows all the constructing nodes, implicit constraints \(\mathcal{N}^{\square}\), polynomials, and durations. 
In addition, a constraint \(\mathcal{C}_k\), discretized at a fixed time step \(k\), is shown; such a type of constraint can be used to enforce the dynamics equality or the range-of-motion inequality constraints.
Notably, these can be enforced not only at the nodes but anywhere along the polynomials.
The optimization values that solely represent values for a period of time \(t\) are denoted by \(\mathcal{N}^{-}\).

In this work, we use the SRBD model
and employ geometric shapes known as \emph{superquadrics},
shown in Fig.~\ref{fig:exp_lmtr_snapshots},
as range-of-motion constraints
described by:
\begin{equation}
\lvert p_{i,x}(t)/A \rvert^a + \lvert p_{i,y}(t)/B \rvert^b + \lvert p_{i,z}(t)/C \rvert^c = 1
\end{equation}
whose curvature is defined by the exponents \({a, b, c}\) and dimensions are specified by the scalings \({A, B, C}\).
This shape conveys a more natural range of foothold choices
and avoids unwanted behaviors that a box shape can promote, \ie, yawing to maximize the size of a stride.

%% file: framework_heuristic.tex
\subsection{Heuristic Initialization} \label{sec:heuristic_initialization}
The \iH{} initializes the base trajectory variables using the corresponding values of the solution obtained at the previous replanning cycle.
The non-overlapping part of the new base trajectory is linearly interpolated up to the user-defined goal.
The footholds corresponding to the first stance phase coincide with the location given by the previous plan whilst the footholds of the second stance phase are located at a nominal distance from target goal.

Unlike LMTR, the Heuristic initializes the NLP with periodic gaits which result in the two stance phases having the same duration ($\Delta T_{i,\text{st1}} =\Delta T_{i,\text{st2}} =\Delta T_{i,\text{st}}$) and in all the feet having the same stance and swing durations $\Delta T_{\text{st}}$ and $\Delta T_{\text{sw}}$ (these values can, however, still be varied during the optimization).
The durations
can be set using the duty factor $d$ and the cycle factor $c$ such that $\Delta T_{\text{st}} + \Delta T_{\text{sw}} = (1.0 - c) \Delta T_H$ and $d = \Delta T_{\text{st}} / T_{\text{c}}$ (where $T_{\text{c}}$ is the gait cycle time $T_{\text{c}} = \Delta T_{\text{st}} + \Delta T_{\text{sw}}$).
The relative time offset $t_{0,i}$ between the start of the base trajectory and that
of the $i^\text{th}$ foot can be set as the fraction $\alpha_i \in (0,1)$ such that: $ t_{0,i} = \alpha_i \Delta T_H$.

%% file: framework_lmtr.tex
\subsection{Learned Initialization via \iL{}} \label{sec:framework_lmtr}
Solving a complex, long-horizon NLP \emph{from scratch} can require many seconds of computation per second of generated motion.
Using shorter horizons reduces computation but makes performance increasingly limited 
by how intelligently the final boundary conditions can be defined.
We approach these issues via imitation learning of a set long-horizon expert trajectories
that have been processed into segments, as originally detailed in~\cite{surovikLearningExpertSkillSpace}.
This produces a regression model (\iL{}) that expresses a diverse continuum of behaviors
suited for handling a range of obstacle configurations.

\subsubsection{Change of Scope}
\begin{figure} [t]
    \centering
    \subfigure[Stationary Horizon (SH)]{
    \includegraphics[height=17mm,trim={0mm 15mm  0mm 15mm},clip=true]{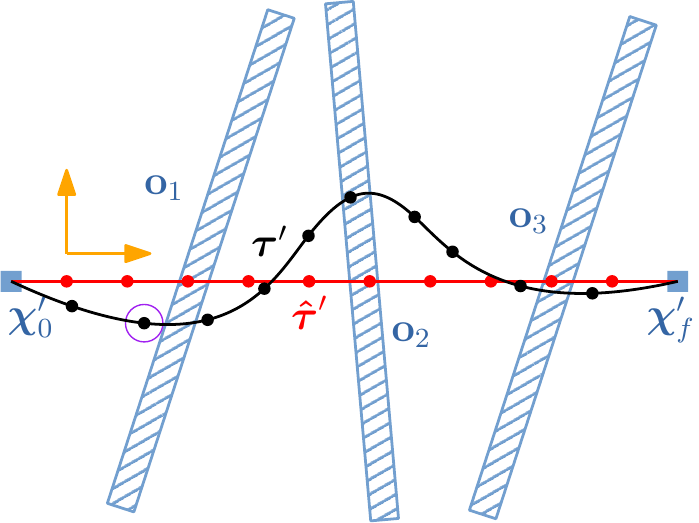}
}
\hspace{-5mm}
    \subfigure[Receding Horizon (RH)]{
        $\;$
    \includegraphics[height=17mm,trim={6mm 15mm 25mm 15mm},clip=true]{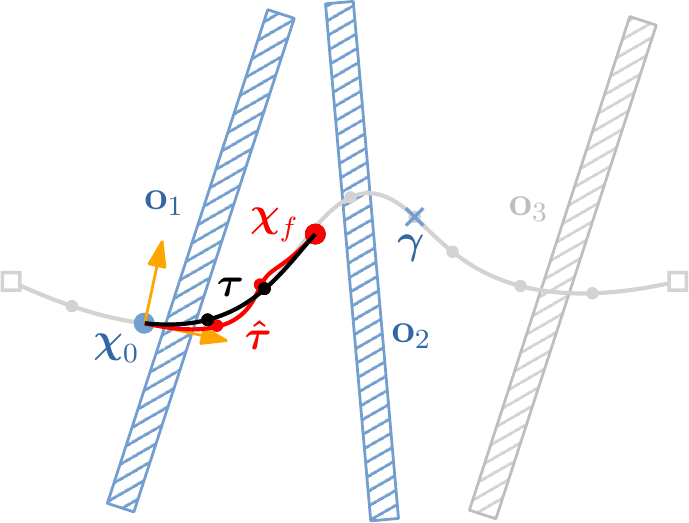}
}
	\caption{
        Illustration of two scopes with which to view trajectories 
        that cross discrete obstacles (rectangles).
        Blue indicates components of the ``task'', \ie, inputs.
        Red and black denote outputs of the initializer and optimizer, respectively.
        (a) SH scope, with static robot states at
        the boundaries of a long trajectory, and naive initialization.
        (b) RH scope, with a trajectory segment
        bounded by dynamic states, and learning-based initialization (\iL{}).
	}
	\label{fig:scopes}
\end{figure}

Key to the imitation learning setup is the use of segmentation (Sec.~\ref{sec:problem_formulation})
to convert expert data from \emph{stationary-horizon} (SH) to \emph{receding-horizon} (RH) scope,
as sketched in Fig.~\ref{fig:scopes}, 
with SH variables using $\stathoriz\cdot$ notation.
SH involves static boundary states $\stathoriz\state$ and long trajectories $\stathoriz\traj$,
while in RH $\state$ are dynamic and $\traj$ are shorter segments.
Initialized values are denoted $\hat\cdot$, 
and $\terrline_i$ are discrete terrain elements.
In RH scope, the task
$\task = \left(\state_0, \goal, \terrline_{i},\terrline_{i+1}\right)$
includes $\terrline_i$ within a local window and an SE(2) goal $\goal$ taken from a base pose on the SH trajectory.
With $\goal$ occurring further along than $\state_f$,
the size of $\traj$ can be controlled separately from the
amount of foresight encompassed by $\task$.

\subsubsection{Expert Dataset}
\begin{figure} [b]
	\includegraphics[width=\columnwidth]{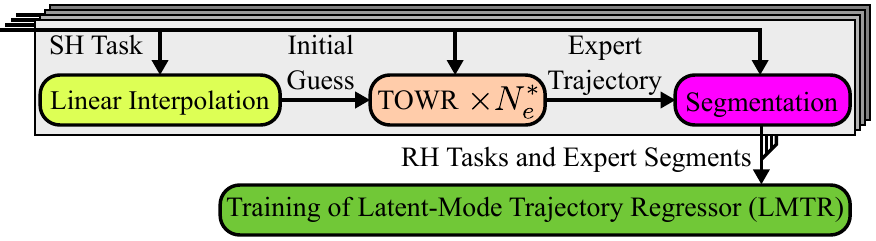}
    \vspace{-1.5em}
	\caption{
        Diagram of bulk data generation for imitation learning.
        Stationary-horizon tasks are sampled and extensively optimized into high-quality trajectories.
        The segmentation process converts data into receding-horizon segments and tasks,
        used for training the multi-modal trajectory regressor.
	}
	\label{fig:learning}
\end{figure}

Training data are generated in bulk as in Fig.~\ref{fig:learning}.
Within SH scope, tasks $\stathoriz\task$ are sampled from a distribution of static robot states and terrains
and $\stathoriz{\hat\traj}$ are initialized naively with linear interpolation and standard gait timings.
Allowed a high number $N_{max}$ of iterations per sample, 
TOWR with analytical costs \cite{melonReliableTrajectoriesDynamic2020a} returns 
expert trajectories $\stathoriz\traj$.
The segmentation process of Sec.~\ref{sec:problem_formulation} is then used to generate the RH dataset,
\ie, pairs of RH tasks and segments $\left\{\task,\traj\right\}$,
desired for imitation learning.

\subsubsection{Generalization via \iL{}} \label{sec:generalization}
Due to the strong nonconvexity of the NLP,
similar $\task$ may be paired with substantially different $\traj$,
reflecting a rich inventory of behaviors.
This prevents directly fitting a ``unimodal'' regressor $f:\task\longrightarrow\traj$, which must smoothly vary based on $\task$ alone.
Instead, a ``mode'' $\mode$ that represents task-independent variation can be learned in an unsupervised fashion by training a Conditional Variational Autoencoder, where $\traj$ is the reconstructed variable, $\task$ is the condition, and $\mode$ is the latent variable. 
The decoder module of this model then serves as the regressor $f:\left(\mode;\task\right)\longrightarrow\traj$ deployed online.
Selection of $\mode$ is done using nearest-neighbor lookup on expert pairs $\left\{\task,\mode\right\}$
with a weighted distance metric.

%% file: framework_integration.tex
\subsection{State Prediction due to Optimization Delay} \label{sec:state_prediction}
Most MPC algorithms target relatively high frequencies of \SIrange{40}{200}{Hz} \cite{bledtImplementingRegularizedPredictive2019, bledtExtractingLeggedLocomotion2020}.
Because our framework is concerned with solving longer-horizon dynamic motion, it operates at a much lower, fixed frequency of a few \SI{}{Hz}.
As a result, it experiences delays measured in hundreds of milliseconds between measurements of the state and execution of resulting solutions.
To manage this issue, the starting point of the computed trajectory corresponds to the time at which it will be executed and occurs after the computation begins.
We implement a simple \emph{predictor} to handle this by using the desired position computed by the previous plan and the current tracking error:
\begin{align}
\hat{\statep}(t_{0, k+1}) &= \statep^{*}_{k}(t_{0, k+1}) + \boldsymbol{\alpha}(t) \odot \overline{\statep}(t) \\
\overline{\statep}(t) &= \statep(t) - \statep^{*}_{k}(t)
\end{align}
Where $t$ is the current time, \(t_{0, k+1}\) is the time at start of the new plan \(k+1\), \(\hat{\statep}(t_{0, k+1})\) is the predicted state used as initial state for the next plan, \(\statep^{*}_{k}(t)\) is the desired state from the current plan, \(\overline{\statep}(t)\) is the current tracking error and \(\boldsymbol{\alpha}\) is a scaling vector.

For the scaling, values of $0$ means that the prediction relies only on the desired state while values of $1$ fully take into account the tracking error and correspond to the actual state of the robot when prediction is with a short horizon.
For each foot, the scaling vector is set to $\mathbf{1}$ if it is in stance and this phase corresponds to the same one at the beginning of the next plan, or $\mathbf{0}$ otherwise.
For the base, the scaling factors are tuned between those two values to allow feedback on the actual state while still being close enough to the previous plan for stable numerical behavior. 
This kind of simple prediction is only valid if the horizon is short with respect to the dynamics of the tracking error.
To have a more accurate initial state for the optimized trajectory, the constraint on the initial state is updated with a new prediction between each iteration of the optimizer.

\subsection{Terrain Perception} \label{sec:terrain_perception}
The onboard RealSense depth cameras and the GridMap library~\cite{fankhauserUniversalGridMap2016} are used to perceive the terrain. 
GridMap constructs a 2.5D heightmap which constrains the footholds and a costmap to aid obstacle avoidance.
A raw version of the heightmap is used to match the $z$ coordinate of the planned footholds to the actual terrain and a smooth version of the heightmap is used to set a cost on each candidate foothold.
By making the cost proportional to the slope of the terrain,
the solver is incentivized to choose footholds away from edges and inclined surfaces.
In addition, the raw heightmap data is processed to detect edges which become part of the task input provided to the \iL{}.
Both the heightmap and the edges are updated at a frequency of \SI{6}{Hz}.

\subsection{Whole-Body Controller and Robustness to Terrain} \label{sec:whole_body_control}

To execute the generated plans, the whole-body
controller~\cite{dariobellicosoPerceptionlessTerrainAdaptation2016}
is used to track the
trajectory of the base and the end-effectors at 400 Hz.
The controller includes different heuristics to improve its behavior in case of slip or other unexpected events.

To be more robust to errors in the terrain height measured by the depth cameras and to be able to make contact at the right instant, the swing trajectory has been modified.
On the planner side, the formulation of splines in TOWR has been adapted to allow non-zero vertical velocities for the foot motions at the transition from swing to stance.
Additionally, the controller is able to modify the setpoints of the feet computed by the planner by continuing the swing trajectory downward until contact is detected.
Those two features allow our controller to quickly react to delayed contacts while having smooth foot trajectories despite the low frequency of the proposed planner.

%% file: results.tex
\section{Results and Discussion} \label{sec:results}

The receding-horizon planner was tested with both trajectory initialization options, \iH{} and \iL{}.
The robot was tasked with computing dynamic motions toward a goal. The quality of the initial guesses and the refined trajectories were quantified by analyzing the cost and constraint violation computed at each iteration of the nonlinear solver.
Execution of the replanning pipeline on the ANYmal C quadruped demonstrated its viability on hardware for most scenarios.

In the most challenging case---aggressive behaviors on terrain planned by \iL{}---viability on hardware 
or in high-fidelity physics simulators
has not been fully achieved. 
Instead, we provide numerical trials in which the replanner receives the task that would be seen if the previous plan were followed nominally for the duration of one replanning cycle, and refinement is conducted using the same computation time limit as hardware trials.

\begin{figure} [t!]
    \includegraphics[width=\columnwidth]{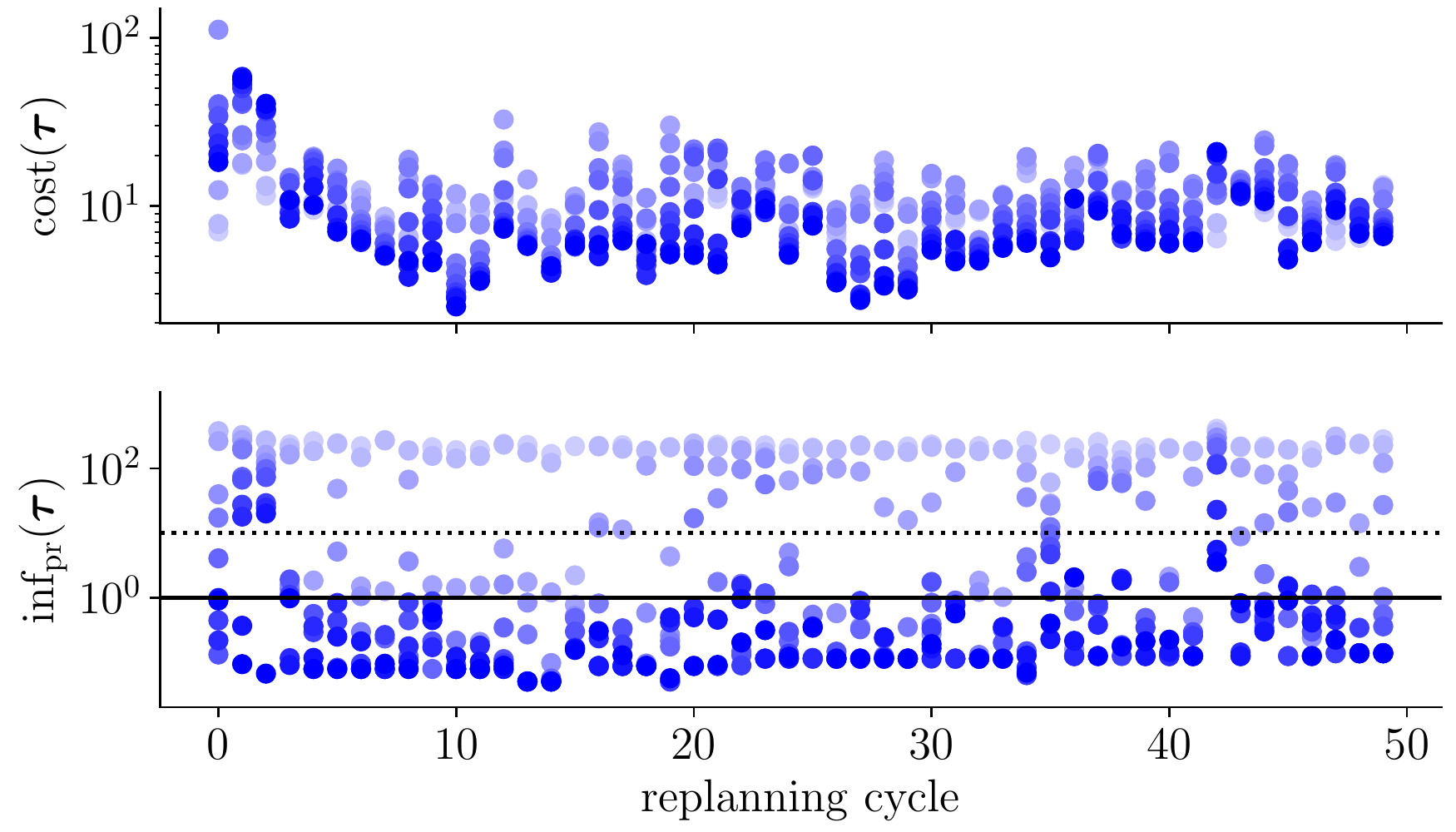}
    \vspace{-1em}
    \caption{
        The values of \(\mathrm{cost}(\primals)\) and \(\mathrm{inf_{pr}}(\primals)\)
        during refinement of initial guesses provided by \iL{} while traversing the terrain shown in Fig.~\ref{fig:exp_lmtr_snapshots} for \SI{50}{} replanning cycles.
        The gradient (light to dark) represents subsequent iterations limited to a maximum of \SI{10}{} per cycle.
        Horizontal lines show preferred (solid) and acceptable (dotted) thresholds of \(\mathrm{inf_{pr}}(\primals)\).}
    \vspace{-0.5em}
    \label{fig:refinement}
\end{figure}
\begin{figure} [b!]
    \vspace{-1em}
    \includegraphics[width=\columnwidth]{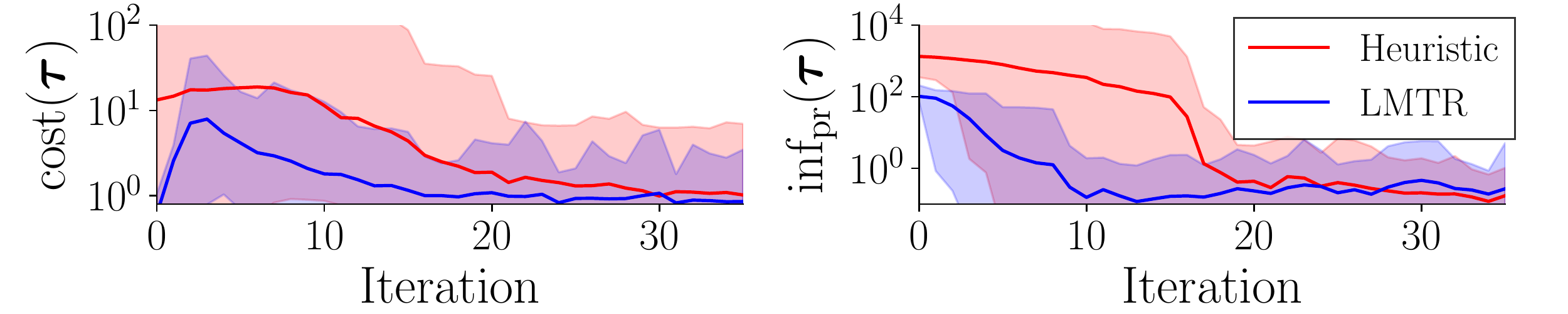}
    \vspace{-2em}
    \caption{
        Mean values of the cost and primal infeasibility at each refinement iteration of \SI{1.2}{\second} long initial guesses provided by the \iH{} (linear interpolation) and \iL{} (near-optimal trajectories) on flat terrain at \SI{2}{\hertz}.  
        Here, both cases use the same mid-motion final state $\state_f$
        intelligently provided by \iL{} as shown in Fig.~\ref{fig:scopes}% 
        ---a key strength of the regressor.
        }
    \label{fig:refinement_comparison}
\end{figure}

\begin{figure*} [t!]
    \includegraphics[width=\textwidth]{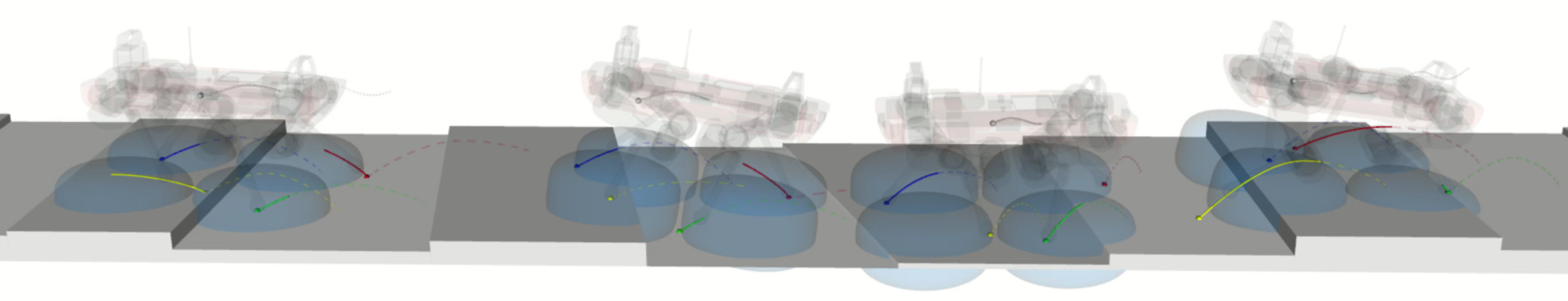}
    \vspace{-0.7cm}
    \caption{
        Snapshots of a typical numerical trial of the \iL{}
        which leverages similar prior experience
        to quickly plan very dynamic trajectories in a receding-horizon fashion 
        while being informed by real-time perception.
        Steps were of arbitrary depth, height and orientation.
        The generated plans were refined online by the solver.  
        The dashed lines indicate planned base and feet trajectories while the solid lines emphasize the portion of the plan executed during each MPC control cycle. 
        The blue superquadrics approximate the operational space of the end-effectors by serving as range-of-motion constraints.
    }
    \label{fig:exp_lmtr_snapshots}
\end{figure*}
\begin{figure} [b!]
    \vspace{-1em}
	\includegraphics[width=\columnwidth]{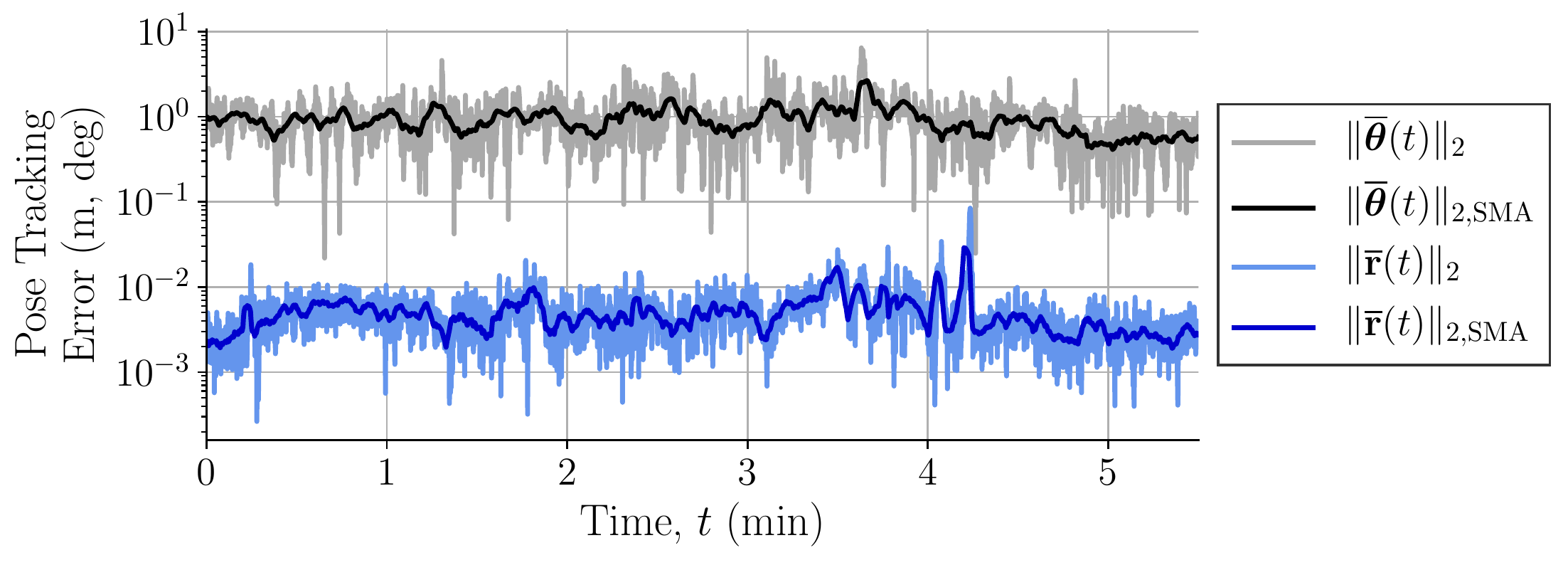}
    \vspace{-2em}
	\caption{
        Despite a low replanning frequency (\SI{1.4}{\hertz}) 
        and a large time step \(dt_{\text{base-poly}}\) (\SI{0.1}{\second}),
        the direct collocation formulation produces trajectories which satisfy the dynamics to third-order
        enabling accurate and precise tracking by the whole-body controller.
        The figure shows base pose tracking errors,
        and their simple moving averages (SMA), 
        or a \SI{5.5}{\minute} execution of the MPC initialized by the \iH{},
        on terrain, by the real robot.
        The root-mean-squared error (RMSE) values of the linear position and orientation were \SI{6.55e-3}{\metre} and \SI{1.07}{\degree}, respectively. 
    }
    \vspace{-0.5em}
    \label{fig:base_tracking_error}
\end{figure}

\subsection{\iL{} Setup and Numerical Performance} \label{sec:reconstruction_performance}
An \iL{} regressor was trained for both environment types, 
using 
$\mode\in\mathbb{R}^{12}$ and with
$\traj\in\mathbb{R}^{312}$ containing a \SI{1.2}{\second} base horizon and stance-swing-stance phases.
For the \emph{flat-ground} scenario, the offline expert's initial conditions used random offsets of the lateral position and the yaw relative to a desired line of travel. 
Without any obstacles, $\task\in\mathbb{R}^{27}$.

The \emph{terrain} scenario included randomized stairstep features $\terrline_i$ spaced apart by \SIrange{0.25}{1}{\metre}, with height changes of \SIrange{-0.16}{0.16}{\metre} and edge orientations of \SIrange{-15}{15}{\degree}.
During deployment, these terrain features are extracted from the elevation map,
and the closest two are included in $\task\in\mathbb{R}^{35}$.
Numerical results shown in Fig.~\ref{fig:exp_lmtr_snapshots} demonstrate motions computed at \SI{2.5}{\hertz}
that reach mathematical feasibility under the SRBD formulation of Sec.~\ref{sec:problem_formulation}%
---attaining reliable deployment in physics simulators and on hardware is a prime focus of our ongoing work.

When facing a new scenario within the bounds of the experience set, the network consistently reconstructed a valid initial guess which was refined to a satisfactory level of optimality within the iteration budget available for online use, as shown in Fig.~\ref{fig:refinement}.
Occasionally, some cycles presented conditions that were challenging to satisfy, potentially due to gaps in the regressor's coverage.
If refinement failed, the remaining portion of the previous valid trajectory was used for an additional replanning cycle, which was possible due to a greater than 2-to-1 ratio of these two durations.

Figure~\ref{fig:refinement_comparison} shows the evolution of the cost and the constraint violation along the iteration of the solver for the \iH{} and the \iL{} initialization. 
While the \iH{} could be used with any predefined gait, to have a fair comparison with respect to the difficulty of the gait, both were evaluated on the contact timing and footholds generated by the regressor. The results show that the \iL{} initialization was able to converge around 2 times faster than the \iH{}.

\subsection{Experimental Evaluation} \label{sec:experimental_evalution}
\paragraph{Heuristic initialization}
The initializations provided by the \iH{}, based on user-commanded target robot base pose, enabled the continuous execution of a walk and a trotting gait during which the robot traversed obstacles of heights up to \SI{0.2}{\metre} as shown in Fig.~\ref{fig:main}.
The base tracking error is shown in Fig.~\ref{fig:base_tracking_error} for a hardware experiment where the robot walked over obstacles up to \SI{0.13}{\metre} high for over \SI{5.5}{\minute}.

The replanning frequency for the walk gait was \SI{1}{\hertz} with an optimization horizon of \SI{3}{\second}.
The trot could run at \SI{2}{\hertz} over a replanning window of \SI{1}{\second}.
For both walk and trot, the Heuristic was operated at speeds of about \SI{0.1}{\metre \per \second} on flat ground and while negotiating the obstacles shown in the accompanying video.
The average step length on flat ground was \SI{0.25}{\metre} for the trot and \SI{0.3}{\metre} for the walk.

\begin{figure} [t!]
    \includegraphics[width=\columnwidth]{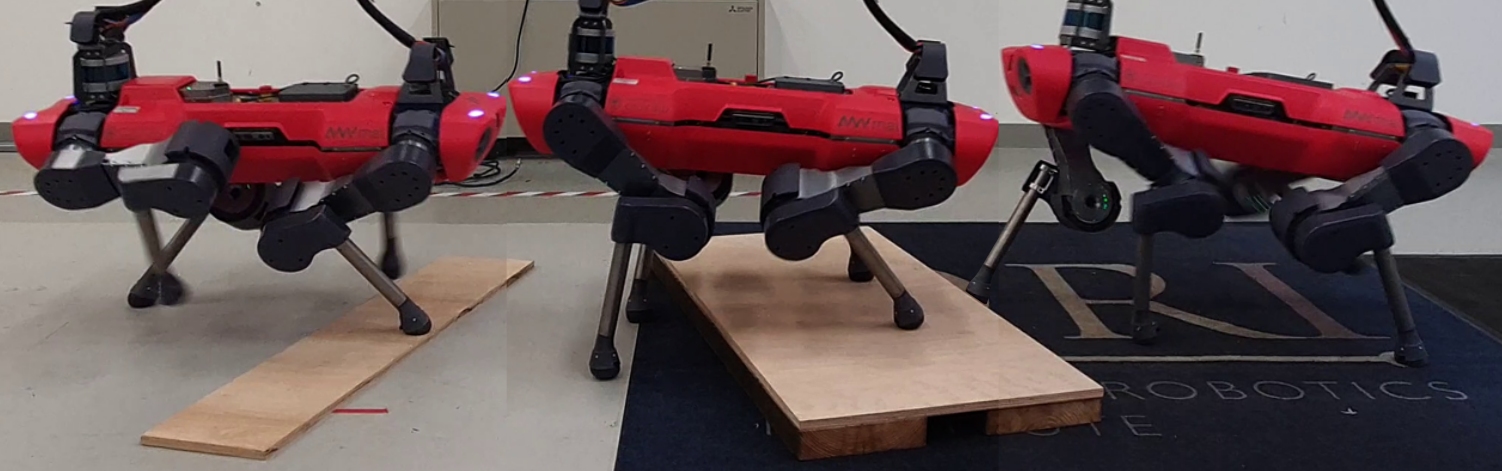}
    \caption{        
        The initial guesses provided by the \iL{} trained on flat-ground
        can be adapted to mild height variation of \SIrange{0.02}{0.08}{\metre}
        by the receding-horizon optimization,
        showing the ability to cope with minor imprecisions.
    }
    \label{fig:exp_nn_terr}
\end{figure}

\paragraph{LMTR initialization}
In the experiments with the \emph{learned} initialization, the \iL{} was used to enable a \SI{4}{\metre}, \SI{12}{\second} trot-like motion on flat ground which resulted in the base velocity in the \(x\)-direction \(\dot{r}_{x}(t)\) reaching \SI{0.7}{\metre\per\second} and swing lengths of \SIrange{0.2}{0.6}{\metre} with durations \(\Delta \mathbf{T}_{\text{sw}}\) of \SIrange{0.3}{0.5}{\second} .
Thanks to the \iL{} initialization, such a plan can be adapted online by the optimizer at frequencies from \SIrange{2}{3.3}{\hertz} with planning horizons of \SI{1.2}{\second} to take into account the model of the terrain received from the sensors and walk over obstacles as shown in Fig.~\ref{fig:exp_nn_terr}.

%% file: conclusion.tex
\section{Conclusion} \label{sec:conclusion}
This work presents a pipeline for iterative perception-aware planning of dynamic legged locomotion in a receding-horizon fashion via online trajectory optimization with a novel asynchronous, segmentation-based formulation using the single rigid body dynamics (SRBD) model. 
We present two initialization schemes.
The \emph{\iH{}} combines previous solutions with a linearly interpolated guess to accelerate the computation.
The \emph{latent-mode trajectory regressor~(\iL{})}, which imitates expert data, is used to extend the planning horizon, proposing a behavior dependent on the task at hand, which may include prominent terrain features.
We deploy this pipeline on a quadrupedal robot, ANYbotics ANYmal C, using fully onboard sensing and computation.
Hardware trials with the \iH{} demonstrate the ability to traverse terrain obstacles using walking and trotting gaits.
The \iL{} enabled more dynamic, trot-like locomotion but with aperiodic contact sequences that hint at its capacity to express a diverse inventory of viable behaviors.
Numerical results show how its expressiveness is suitable for adapting to challenging terrain while moving at fast pace.
Future work will seek to realize these precise and highly dynamic motions by explicitly compensating for the approximations in the planning model, in lower-level control, and by updating the \iL{} based on deployed experiences.